\documentclass[lettersize,journal]{IEEEtran}
\usepackage{amsmath,amsfonts}
\usepackage{algorithmic}
\usepackage{algorithm}
\usepackage{array}
\usepackage[caption=false,font=normalsize,labelfont=sf,textfont=sf]{subfig}
\usepackage{textcomp}
\usepackage{stfloats}
\usepackage{url}
\usepackage{color}
\usepackage{verbatim}
\usepackage{graphicx}
\usepackage{cite}
\usepackage{amssymb}
\usepackage{mathtools}
\usepackage{amsthm}
\usepackage{multirow}
\usepackage{pifont}

\begin{document}

\title{Knowledge-guided Complex Diffusion Model for PolSAR Image Classification in Contourlet Domain}

\author{Junfei Shi, ~\IEEEmembership{Senior Member, IEEE}, Yu Cheng, Haiyan Jin, ~\IEEEmembership{Member, IEEE}, Junhuai Li,~\IEEEmembership{Member, IEEE}, Zhaolin Xiao, ~\IEEEmembership{Member, IEEE}, Maoguo Gong,~\IEEEmembership{Fellow, IEEE}, Weisi Lin, ~\IEEEmembership{Fellow, IEEE} 

\thanks{Junfei Shi, Yu Cheng, Haiyan Jin, Junhuai Li and Zhaolin Xiao are with the Department of Computer Science and Technology, Shaanxi Key Laboratory for Network Computing and Security Technology, Xi'an University of Technology, Xi'an 710048, China. Corresponding author: Junfei Shi.(email:shijunfei@xaut.edu.cn.)}
\thanks{Maoguo Gong is with School of Computer Science and Technology, China University of Mining and Technology, Xuzhou, China.}
\thanks{Weisi Lin was with the School of Computer Science and Engineering, Nanyang Technological University, Singapore 639798.}
}

\markboth{Journal of \LaTeX\ Class Files,~Vol.~14, No.~8, May~2024}%
{Shell \MakeLowercase{\textit{et al.}}: A Sample Article Using IEEEtran.cls for IEEE Journals}


\maketitle

\begin{abstract}
Diffusion models have demonstrated exceptional performance across various domains due to their ability to model and generate complicated data distributions. However, when applied to PolSAR data, traditional real-valued diffusion models face challenges in capturing complex-valued phase information. Moreover, these models often struggle to preserve fine structural details. To address these limitations, we leverage the Contourlet transform, which provides rich multiscale and multidirectional representations well-suited for PolSAR imagery. We propose a structural knowledge-guided complex diffusion model for PolSAR image classification in the Contourlet domain. Specifically, the complex Contourlet transform is first applied to decompose the data into low- and high-frequency subbands, enabling the extraction of statistical and boundary features. A knowledge-guided complex diffusion network is then designed to model the statistical properties of the low-frequency components. During the process, structural information from high-frequency coefficients is utilized to guide the diffusion process, improving edge preservation. Furthermore, multiscale and multidirectional high-frequency features are jointly learned to further boost classification accuracy. Experimental results on three real-world PolSAR datasets demonstrate that our approach surpasses state-of-the-art methods, particularly in preserving edge details and maintaining region homogeneity in complex terrain.
\end{abstract}

\begin{IEEEkeywords}
PolSAR image classification, Diffusion model, Remote sensing, Contourlet domain, Knowledge guide.
\end{IEEEkeywords}

\section{Introduction}
\IEEEPARstart{P}{olarimetric} Synthetic Aperture Radar (PolSAR) system can obtain polarimetric scattering echoes under all weather conditions and throughout the day and night. Unlike optical and infrared images, PolSAR data capture scattering echos from multiple polarimetric modes, which provides richer insights into the scattering mechanisms of various objects on the Earth's surface. PolSAR images play a crucial role in remote sensing image applications, offering unique advantages in the analysis of surface structures, vegetation, water bodies, and urban areas. As a result, PolSAR has gained significant importance in fields such as environmental monitoring\cite{RN85}, land cover classification\cite{RN84}, change detection\cite{RN87}, and agricultural evaluation\cite{RN86}.

Traditional PolSAR image processing techniques consist mainly of target decomposition, statistical distribution and machine learning methods, which have demonstrated advantages in image classification.  Target decomposition-based methods include Huyen and Cloude\cite{RN15}, Pauli decomposition\cite{RN30}, Freeman and Durden\cite{RN16}, Yamaguchi decomposition\cite{RN17}, etc.  These methods can obtain rich polarimetric information to reflect the scattering mechanism of terrain objects. In addition, statistical distribution-based methods have received much attention from researchers, since PolSAR data have extremely complicated distribution characteristics, especially for heterogeneous regions.  Commonly used distributions include Wishart\cite{RN32}, G0\cite{RN34}, KummerU\cite{RN33}, etc.  Later, machine learning methods have been introduced to PolSAR classification, such as SVM\cite{RN78}, fuzzy cluster\cite{RN79}, MRF\cite{RN80},etc. However, these methods rely on hand-crafted features, which are often limited in their ability to capture complex, high-dimensional semantic information, especially for heterogeneous terrain objects.

In recent years, deep learning models have shown remarkable performance in extracting semantic features and capturing intricate patterns of PolSAR images. Mainstream models includes CNN, Transformer etc. CNN-based methods aim to capture spatial and contextual dependencies, as evidenced by models such as PolMPCNN\cite{RN48},SF-CNN\cite{RN53},MP-ResNet\cite{RN38} and 3D-CNN\cite{RN54}. Recently, Transformers have attracted increasing attention owing to their capacity to model global dependencies via self-attention mechanisms, such as TCSPANet\cite{RN37} and HybridCVNet\cite{RN55}. Considering PolSAR complex data, many complex-valued methods have been proposed. For instance, CV-CNN\cite{RN50} introduces complex-valued convolutional operations to directly process complex-valued input, CV-CNN-SE\cite{RN56} integrates polarimetric feature extraction with squeeze-and-excitation (SE) modules to enhance PolSAR image classification performance with minimal computational overhead. All of these methods can learn high-level semantic information, producing good performance for complicated objects. However, these models still face challenges in fully exploiting the unique characteristics of PolSAR images, such as their inherent multi-resolution and statistical characteristics.

Diffusion model, also known as a robust and stable generative model \cite{RN20,RN57} has achieved impressive results in various domains, including natural language processing\cite{RN40,RN88}, time-series forecasting \cite{RN41,RN52}, etc. A typical diffusion model comprises a forward process that gradually perturbs the input by adding Gaussian noise, and a reverse process that reconstructs the original data via iterative denoising. In computer vision, this process involves training a neural network to learn the reverse diffusion trajectory, effectively denoising images corrupted by Gaussian noise \cite{RN58}. Diffusion models have recently attracted significant attention due to their flexibility and robustness. They have been successfully applied to a wide range of complex visual tasks, including image generation \cite{RN43,RN57,RN20}, image fusion~ \cite{RN58}, and image super-resolution \cite{RN59,RN94,RN93}. For image classification tasks, diffusion models, such as SBGC \cite{RN43},\cite{RN89} and SpectralDiff \cite{RN44}, effectively capture underlying patterns and dependencies by modeling the data distribution, thereby achieving promising performance.

Although diffusion models have demonstrated success in certain image processing tasks, they still face limitations when dealing with complex PolSAR data:

\begin{itemize}
	\item Fail to learn complex-valued data structure: Diffusion models generally convert complex-valued PolSAR data to real vector, which apparently destroy the complex structures and ignore the phase information. So, diffusion model with real-valued Guassian noises often struggle to capture high-dimensional complex data distribution, resulting in suboptimal representation of PolSAR data. Therefore, complex-valued diffusion model is necessary to be exploited.
	\item Ignore edge structural information: One of the critical challenges in PolSAR image classification is the preservation of edge details, which are essential for accurate interpretation. However, diffusion models tend to learn statistical textural information during the iterative denoising process, resulting in the smooth or loss of important structural details. Therefore, diffusion model-based methods are difficult to learn both statistical and structural information simultaneously. Effectively integrating statistical textures and structural edges into a unified framework remains a significant challenge.
\end{itemize}

Contourlet transform is an effective multi-resolution analysis method. Compared to wavelets, it captures richer multi-scale and multi-directional geometric features, making it particularly suitable for PolSAR image analysis. Low-frequency signals represent statistical texture information, while high-frequency subbands capture multidirectional structural details. This separation makes it particularly suitable for simultaneously learning both statistical (low-frequency) and structural (high-frequency) features. Recent methods, such as \( C^2N^2 \) \cite{RN26},KGCSL \cite{RN24}, have leveraged contourlet-based representations to improve edge detection, texture classification, and feature fusion in remote sensing tasks.
However, contourlet features are often directly fed into the network without fully considering their individual strengths or the potential interaction effects between them. For better learning complex PolSAR data, it is essential to explore and integrate multi-scale and multi-directional features in a more adaptive and comprehensive manner.

To address the limitations of existing diffusion-based methods and leverage the advantages of the contourlet transform, we propose a knowledge-guided complex contourlet domain diffusion model for PolSAR image classification. Firstly, complex diffusion model is designed to learn the statistical features of low-frequency bands, and integrates high-frequency contourlet coefficients as structural knowledge to guide the diffusion process, enhancing the learning of edge details while preserving statistical features. In addition, high-frequency subbands with multi-scale and multi-directional information are enhanced to improve the classification performance. This approach allows for the simultaneous extraction of both fine-grained edge information in high-frequency and robust statistical features from low-frequency information, offering a more comprehensive framework for PolSAR image analysis. The main contributions are given as follows.

\begin{itemize}
	\item  we proposed a knowledge-guided complex diffusion model in counterlet domain for the first time. It develops a complex-valued diffusion model for PolSAR data, and explores structural prior in high-frequency as the knowledge, enhancing both edge and statistical features.
	\item A unified framework is designed to combine low-frequency complex diffusion model and high-frequency enhancing module together, joint learning statistical characteristic of complex PolSAR data and multi-scale structure information. It is a multi-resolution learning method that can be applied to complicated scenes with diverse scale and directional objects. 
	\item Experiments on three real PolSAR datasets verify that the proposed methods outperform SOTA methods in both edge preservation and region homogeneity, especially for complicated PolSAR scenes with diverse scale and directional objects.
\end{itemize}

The remainder of this paper is organized as follows. Section II provides the preliminaries. The proposed method is described in detail in Section III. Experimental results and analysis are presented in Section IV. Finally, Section V concludes the paper.

\section{Preliminary}

\subsection{PolSAR Data}

Polarimetric Synthetic Aperture Radar (PolSAR) systems are capable of capturing richer scattering information than traditional single-polarization SAR systems. The backscattered signal in a full PolSAR system can be represented by the scattering matrix $\textbf{S}$:

\begin{equation}
\textbf{S} = 
\begin{bmatrix}
S_{hh} & S_{hv} \\
S_{vh} & S_{vv}
\end{bmatrix}
\label{eq:scattering_matrix}
\end{equation}
where $h$ and $v$ denote the horizontal and vertical polarization modes, respectively. Under the reciprocity condition, it holds that $S_{hv} = S_{vh}$. Under the Pauli basis,  the scattering matrix can be vectored as:
\begin{equation}
\mathbf{\textit{k}} = \frac{1}{\sqrt{2}} 
\begin{bmatrix}
S_{hh} + S_{vv},\ 
S_{hh} - S_{vv},\ 
2 S_{hv}
\end{bmatrix}^{T}
\label{eq:pauli_vector}
\end{equation}

To reduce speckle effect, a multi-look processing\cite{RN39} technique is employed on the scattering vector. The commonly used coherency matrix $\mathbf{T}$ is then achieved by:

\begin{equation}
\textbf{T} = \frac{1}{n} \sum_{i=1}^{n} \mathbf{\textit{k}}_i \mathbf{\textit{k}}_i^{H}
\label{eq:coherency_matrix}
\end{equation}
where $n$ is the number of looks, and $(\cdot)^H$ denotes the Hermitian (conjugate transpose) operator. In the coherency matrix, the diagonal elements are real-valued, and the off-diagonal elements are complex-valued.

\subsection{Diffusion Model}

Diffusion models have emerged as a powerful approach for image generation, relying on a denoising process that incrementally adds Gaussian noise to an image over $t$ steps. This forward process simulates the degradation of the image, while the model learns to reverse it by progressively removing noise and reconstructing the original image. By modeling the noise distributions at each step, diffusion models can generate high-quality images with remarkable detail and realism.

In the forward diffusion process, the noise is randomly sampled from a standard Gaussian distribution $N(0,I)$. The noisy image at time step $t$ is obtained by iteratively adding noise to the original image over $t$ steps. In contrast, during the reverse process of a diffusion model, the generation of an image involves a gradual denoising procedure. Starting from a sample of pure noise, the model predicts and removes noise at each time step, progressively refining the image through a series of steps. The training of the diffusion model is guided by a loss function, which is defined as follows:
\begin{equation}
\scalebox{0.97}{%
$L=E_{t\sim[1:T],x_0,\varepsilon_t}\left[\left\|\varepsilon_t-\varepsilon_\theta\left(\sqrt{\overline{\alpha}_t}x_0+\sqrt{1-\overline{\alpha}_t}\varepsilon_t,t\right)\right\|^2\right]$%
}
\end{equation}
where $\alpha$ serves as a hyperparameter that controls the magnitude of noise added at each step, while $\varepsilon$ represents the sampling noise, drawn from a Gaussian distribution.


\section{Proposed method}

\begin{figure*} 
    \centering 
    \includegraphics[scale=1.4]{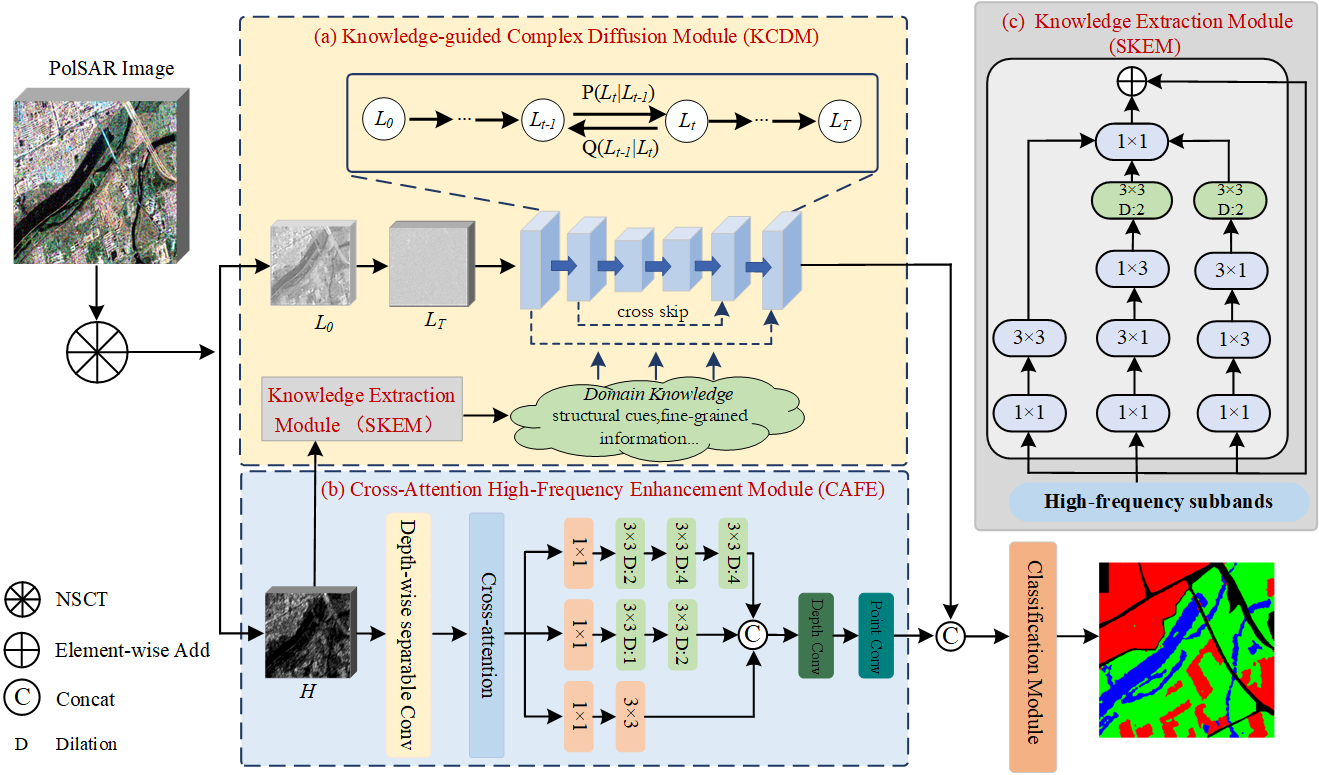} 
    \caption{Framework of the proposed Knowledge-guided complex contourlet diffusion model.} 
    \label{fig:fig1} 
\end{figure*}

Diffusion models are effective in capturing spatial statistical features from PolSAR images but often overlook fine edge details. To address this limitation, we propose a Knowledge-guided Complex Diffusion Model in the Contourlet Domain (CD-KCDM) for PolSAR image classification. As illustrated in Fig.~\ref{fig:fig1}, the CD-KCDM framework begins with a complex-valued contourlet transform that extracts multi-scale and multi-directional coefficients, preserving both low-frequency textures and high-frequency edge details. Leveraging this representation, the Knowledge-guided Complex Diffusion Model (KCDM) is designed to learn statistical features from the low-frequency components while incorporating high-frequency edge information as prior knowledge to guide the diffusion process. To further enhance edge representation, a Cross-Attention High-Frequency Feature Enhancement Module (CAFE) is introduced. Finally, features from KCDM and CAFE are fused and passed to a classification module for the final prediction.

\subsection{Complex-valued Contourlet Transformation}
The Complex-valued Contourlet Transform (CVCT) is an extension of the traditional Contourlet Transform, which is designed to capture multi-directional and multi-scale features for complex images and signals.  By incorporating complex-valued filters, CVCT preserves both magnitude and phase information, making it more effective for PolSAR data processing such as image denoising, edge detection, and texture analysis.  To be specific, contourlet transform is shift-invariant, which decomposes an image using a Laplacian pyramid filter (PFB) and a  directional filter bank (DFB).  First, the image passes through the LP to obtain a lowpass image and a bandpass image.  Then, the bandpass image is further input into the DFB for filtering to obtain multi-directional subbands. Compared with wavelet transformation, Coutourlet can obtain richer multi-scale multi-directional information.

\begin{figure} 
    \centering 
    \includegraphics[width=0.48\textwidth]{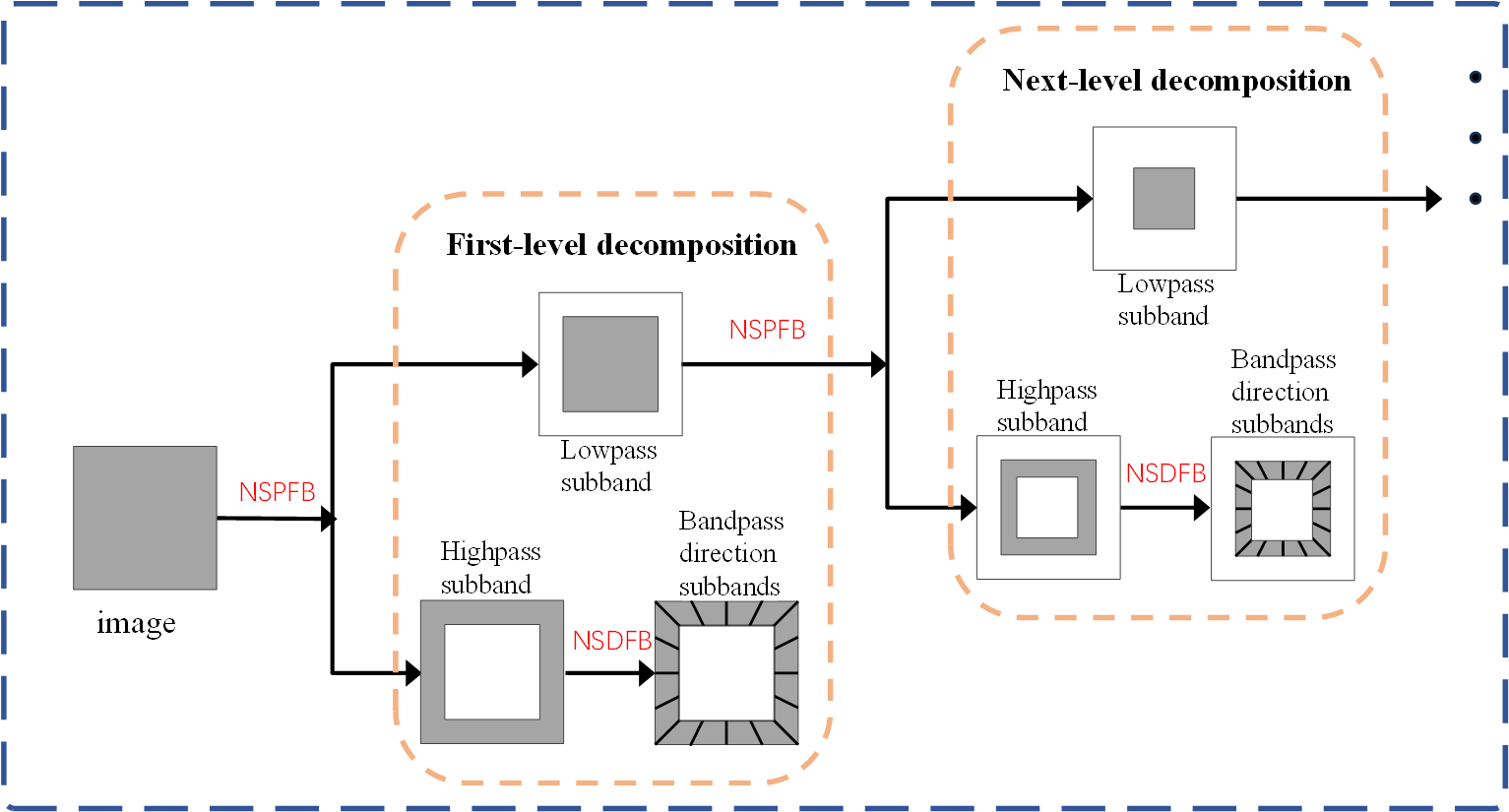} 
    \caption{Non-subsampled contourlet decomposition} 
    \label{fig:contourlet} 
\end{figure}

In this paper, we utilize complex-valued non-subsampled Contourlet transform to extract multi-scale, multi-directional features. As shown in Fig.\ref{fig:contourlet}. The nonsubsampled scheme allows for multi-scale coefficients with the same size, enabling direct utilization of various information for further fusion.  Specifically, for an input X, a set of complex-valued filters are designed for both the non-subsampled Laplacian pyramid filter (NSPFB) and directional filter bank (NSDFB).  After NSPFB filter, an image is decomposed as a low-pass and a band-pass images, defined as:\begin{equation}V_{j-1}=V_j\oplus W_j\quad\mathrm{s.t.}V_{j-1}\supset V_j\end{equation} where \( V_j \) and \( W_j \) represent the lowpass and bandpass images in the \( j \)-th transformation respectively. Then, the bandpass \( W_j \) is then decomposed by a complex NSDFB into \( 2^j \) directional subbands. After the first layer of contourlet decomposition, the complex PolSAR image is split into a low-pass subband and \( 2^j \) high-pass subbands. The lowpass \( V_j \) can be further decomposed in the next layer, resulting in multiscale contourlet coefficients, denoted by
\begin{equation}
C^l = \left\{ X_l^i, \left\{ X_{h,1}^i, \cdots, X_{h,2^j}^i \right\}  \right\}_{i = 0 \sim L}
\end{equation}
where \(2^j\) is the number of directions. \(i\) is the contourlet transformation of the \(i\)-th level, and \(L\) is the total number of levels. The contourlet transformation in the \(i\)-th level can then be expressed as
\begin{equation}
\begin{aligned}
X_{l,i+1} &= X_{l,i} * F_{NSPFB}, \\
X_{h,ds,i+1} &= X_{h,i} * F_{NSDFB},
\end{aligned}
\end{equation}

In this paper, we select \( l=3 \) and \( j=3 \), which means a 3-level complex contourlet decomposition with \( 2^3=8 \) directional subbands at each level. This results in a low-pass subband and high-pass coefficients with 3 scales and 8 directions. Three levels are sufficient to capture multi-resolution and multi-directional information, while too many levels would increase computational complexity, and too few would fail to provide an adequate multi-resolution representation.
\subsection{ Knowledge-guided Complex-valued Diffusion Model in Contourlet Domain}
It is well known that the image texture and most of the energy are concentrated in the low-frequency components. Therefore, statistical features can be learned by developing a knowledge-guided complex diffusion model from low-frequency subbands. The 'knowledge' refers to the geometric structure of the image, which is derived from high-frequency information with multi-scale, multi-directional structural details. This geometric knowledge helps to enhance edge details and guides the diffusion model in learning refined features at the edges. Consequently, two modules are designed in this subsection: the structural knowledge extraction module and the knowledge-guided complex diffusion model.
\subsubsection{Structural knowledge extraction module}
In this module, structural knowledge is extracted as complementary information for low-frequency images. In the low-frequency coefficients, most of the energy is concentrated on capturing the primary textural information of the original image, while fine edge details are often lost. High-frequency coefficients, on the other hand, contain rich detail information. To improve the learning of image statistical features, structural knowledge is derived by combining high-frequency information, which can then guide the learning of low-frequency features. Subsequently, a knowledge-guided complex diffusion model is designed to learn low-frequency statistical features more effectively.

The architecture of the proposed Structural knowledge Enhancement Module (SKEM) is depicted in Fig.~\ref{fig:fig1}(c). This module adopts a three-branch structure based on complex-valued convolutional operations, aiming to capture diverse spatial features at varying receptive fields. The first branch employs a sequence of $1\times1$ and $3\times3$ convolutions to extract basic local features. The second branch introduces asymmetric convolutions in the form of $1\times1$, $1\times3$, and $3\times1$ kernels, followed by a $3\times3$ atrous convolution to enlarge the receptive field without sacrificing resolution. By concatenating the outputs from all three branches and applying a 1×1 convolution for channel compression, the module effectively integrates fine-grained and directional structural cues. This fused representation is then combined with the residual connection, which further enhances the model's capacity to learn complex spatial patterns crucial for accurate PolSAR image classification.

\subsubsection{Knowledge-guided complex diffusion model}

Based on the characteristics of low-frequency components, we propose a complex diffusion model with the guide of structural knowledge to effectively learn and enhance low-frequency information. First, we introduce a complex-valued diffusion model designed to extract complex-valued features from low-frequency data. Building upon a conditional complex diffusion framework, KCDM integrates structural knowledge as a guiding condition, enabling the model to capture intricate statistical features inherent in the low-frequency domain. Furthermore, by leveraging edge and detail information from high-frequency subbands, KCDM refines the learned low-frequency features, thereby enhancing the preservation of geometric structures and textural details.

\textbf{\textit{Complex-Valued Diffusion Forward Process:}}
The conditional diffusion probabilistic model is a widely used generative model that progressively adds Gaussian noise to an image and then removes it to reconstruct the original sample under conditional guidance. 

To adapt this process for PolSAR data, we propose a complex-valued conditional diffusion model. Initially, the input is decomposed using a contourlet transformation to separate it into low-frequency texture and high-frequency structural components.

Let the low-frequency complex-valued input be represented as:
\begin{equation}
\mathbf{L}_0 = \text{Re}(\mathbf{L}_0) + j\text{Im}(\mathbf{L}_0)
\label{eq:L0}
\end{equation}
where $\mathbf{L}_\text{Re}$ and $\mathbf{L}_\text{Im}$ denote the real and imaginary parts, respectively. The forward process can be regarded as a Markov chain, in which complex Gaussian noise is gradually added to the initial state $L_0$. Due to the memory-less property of the Markov process, the probability distribution at step $t+1$ depends only on the distribution at step $t$, which can be formulated as:
\begin{equation}
P(\mathbf{L}_{t+1} | \mathbf{L}_t) = \mathcal{CN}(\mathbf{L}_{t+1}; \sqrt{\alpha_t} \mathbf{L}_t, (1 - \alpha_t)\mathbf{I})
\label{eq:forward}
\end{equation}
where $\alpha_t \in [0, 1]$ denotes the noise schedule controlled by the time step $t$, $I$ is the complex identity matrix, and the probability distribution follows a complex Gaussian distribution, which can be defined as:
\begin{equation}
\mathbf{L}_{t+1} = {\text{Re}} (\mathbf{L}_{t+1})+ j{\text{Im}}(\mathbf{L})_{t+1} \sim \mathcal{CN}(0, \sigma_{t+1}^2 I)
\label{eq:noise}
\end{equation}
where ${\text{Re}} (\mathbf{L}_{t+1})$ and ${\text{Im}}(\mathbf{L_{t+1})} $ denote the real and imaginary parts, respectively. They are independently and identically distributed (i.i.d.).

Using the reparameterization, the noisy data at time step $t$ can be computed as:
\begin{align}
\mathbf{L}_t &= \sqrt{\alpha_t} \mathbf{L}_{t-1} + \sqrt{1 - \alpha_t} \, \epsilon_{t-1} \notag \\
&= \sqrt{\alpha_t} \left( \sqrt{\alpha_{t-1}} \mathbf{L}_{t-2} + \sqrt{1 - \alpha_{t-1}} \, \epsilon_{t-2} \right) + \sqrt{1 - \alpha_t} \, \epsilon_{t-1} \notag \\
&\quad \vdots \notag \\
&= \sqrt{\bar{\alpha}_t} \, \mathbf{L}_0 + \sqrt{1 - \bar{\alpha}_t} \, \epsilon_{0} \\
& =\sqrt{\overline{\alpha}_{t}}\left(\textbf{L}_{0r}+i\textbf{L}_{0i}\right)+\sqrt{1-\overline{\alpha}_{t}}\left(\epsilon_{r}+i\epsilon_{i}\right) \\
 & =(\sqrt{\overline{\alpha}_{t}}\textbf{L}_{0r}+\sqrt{1-\overline{\alpha}_{t}}\epsilon_{r})+i(\sqrt{\overline{\alpha}_{t}}\textbf{L}_{0i}+\sqrt{1-\overline{\alpha}_{t}}\epsilon_{i})
\label{eq:lt_reparam}
\end{align}
where $\bar{\alpha}_t = \prod_{i=1}^{t} \alpha_i$. $\epsilon_{0}=\epsilon_{r}+i\epsilon_{i}$ represents the complex Gaussian noise with independent real and imaginary parts sampled from standard normal distributions. 


Therefore, $\mathbf{L}_t$ follows the following complex Gaussian distribution:
\begin{equation}
P(\mathbf{L}_{t} | \mathbf{L}_0) = \mathcal{CN}\left(\mathbf{L}_{t}; \sqrt{\bar{\alpha}_t} \, \mathbf{L}_0, (1 - \bar{\alpha}_t)\mathbf{I} \right)
\label{eq:forward}
\end{equation}

Given the initial input $\mathbf{L}_0$, the accumulated variance $\bar{\alpha}_t$, and the sampled noise $\boldsymbol{\epsilon_{0}}$, the noised image $\mathbf{L}_t$ at time $t$ can be generated directly.

\textbf{\textit{Knowledge-guided reverse process with complex U-Net:}}

In the complex diffusion model, the goal of the reverse process is to gradually recover the original complex data \( \textbf{L}_0 \) from the noisy one. For the \( t \)-th step, the denoising process is how to obtain the noisy image in the previous step from the current image. Here, a complex U-Net network is designed to learn the denoising process in each step. Given the image \( \textbf{L}_t \), this conditional probability of \( \textbf{L}_\text{t-1} \) can be expressed as:
\begin{equation}Q\left(\textbf{L}_{t-1}|\textbf{L}_t\right)=\mathcal{CN}(\textbf{L}_{t-1};\mu,\sigma_{t}^2I)\end{equation}
where \( \mu \) and \( \sigma \) represent the variance and mean, respectively, and can be derived from the variance factors, noted by:
\begin{equation}\sigma_t^2=\frac{1-\overline{\alpha}_{t-1}}{1-\overline{\alpha}_t}(1-\alpha_t)\end{equation}

The high-frequency information extracted from SKEM is used as knowledge to guide the learned network. The complex U-Net work is utilized to learn the mean value of predicted noises. The procedure is defined as follows:
\begin{equation}\begin{aligned}
\mu & =\mu_\theta(\textbf{L}_t,t,\textbf{H}1) \\
 & =\frac{1}{\sqrt{\alpha_t}}(\textbf{L}_t-\frac{1-\alpha_t}{\sqrt{1-\bar{\alpha}_t}}\epsilon_{_\theta}(\textbf{L}_t,t,\textbf{H}1))
\end{aligned}\end{equation}
where \( \textbf{L}_t \) is low-frequency information, \( \textbf{H}_1 \) is structural knowledge as a condition to guide the reverse process.  \(\epsilon_{_\theta}(\cdot,\cdot)\) is the complex U-Net network for denoising.

\begin{figure} 
    \centering 
    \includegraphics[scale=0.6]{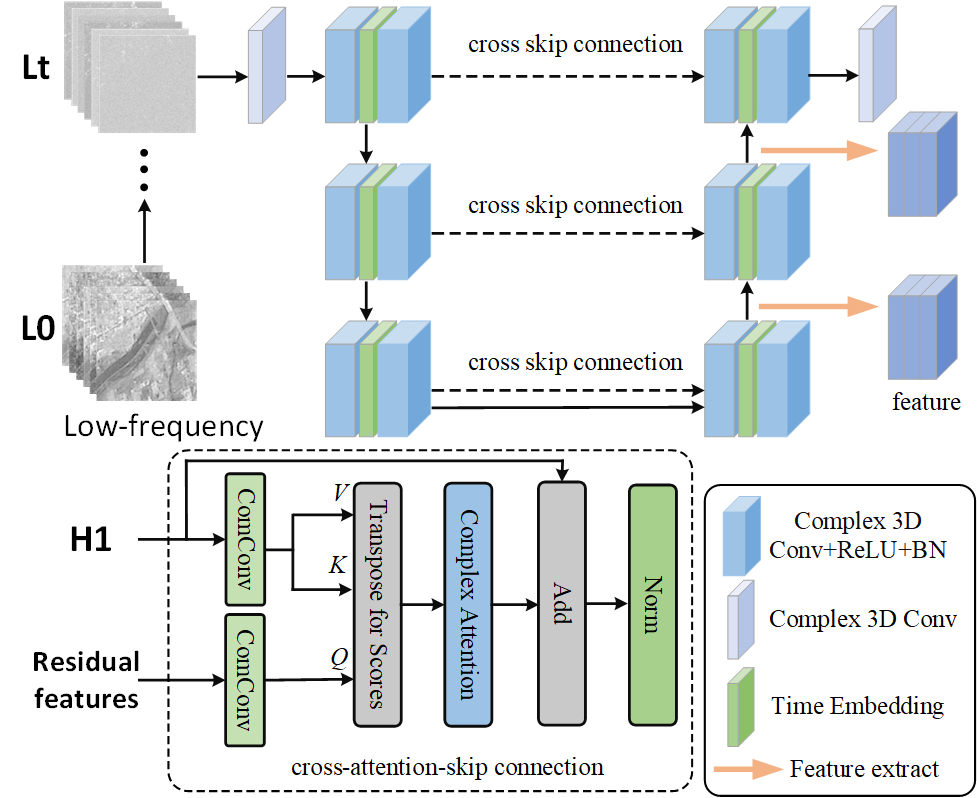} 
    \caption{Network structure of the Knowledge-guided complex U-Net.} 
    \label{fig:unet} 
\end{figure}

The knowledge-guided complex U-Net is shown in Fig.\ref{fig:unet}, which consists of the complex U-Net module and the knowledge-guided cross-attention module. The complex U-Net model can not only extract effective information between channels, but also capture the correlation between the real and imaginary parts at each convolution level. The knowledge guidance module fuses the structural information with residual features at each level of the network through a cross-attention mechanism, which helps the network acquire useful features. 

During the complex U-net model, the complex-valued convolution is utilized to learn the amplitude and phase information. Given the input \( \mathbf{L}_0 \) and the complex weight \( \mathbf{W} = \mathbf{W}_{\text{re}} + j \mathbf{W}_{\text{im}} \), the operation can be expressed as:

\begin{equation}
\begin{aligned}
\text{Re}(\mathbf{Y}) &= \mathbf{W}_{\text{re}} * \text{Re}(\mathbf{L}_0) - \mathbf{W}_{\text{im}} * \text{Im}(\mathbf{L}_0), \\
\text{Im}(\mathbf{Y}) &= \mathbf{W}_{\text{re}} * \text{Im}(\mathbf{L}_0) + \mathbf{W}_{\text{im}} * \text{Re}(\mathbf{L}_0),
\end{aligned}
\label{eq:complex_conv}
\end{equation}

The complex-valued ReLU function is implemented by applying the ReLU activation separately to the real and imaginary parts, which can be expressed as:
\begin{equation}
\text{ReLU}(\mathbf{Y}) = \text{ReLU}(\text{Re}(\mathbf{Y})) + j\,\text{ReLU}(\text{Im}(\mathbf{Y})).
\label{eq:complex_relu}
\end{equation}

The complex-valued 3D convolution operation not only captures the intrinsic correlation between the real and imaginary components, but also effectively extracts informative features across multiple channels. 

To fully exploit structural details and edge information in the images, the network introduces high-frequency guidance at each feature extraction stage. The enhanced high-frequency features are injected into the main network to complement the semantic representations. During upsampling, these high-frequency features are fused with the decoder residual features through a cross-attention mechanism.

During the cross-attention module, the residual features from the U-net network are considered as queries \( \mathbf{Q} \), while the high-frequency guidance features serve as keys \( \mathbf{K} \) and values \( \mathbf{V} \). As shown in Fig.\ref{fig:fig1}(d), after computing the attention scores via the dot product between the query and the conjugate of the key, the real and imaginary parts of the score matrix are separately normalized using the softmax function, resulting in two real-valued attention weight matrices:
\begin{equation}
\mathbf{A}_{\text{Re}} = \text{softmax}(\text{Re}(\mathbf{S})), \quad 
\mathbf{A}_{\text{Im}} = \text{softmax}(\text{Im}(\mathbf{S}))
\label{eq:complex_attn_weights}
\end{equation}
where \( \mathbf{S} = \mathbf{Q} \cdot \mathbf{K}^H \in\mathbb{C}\) is the complex attention score matrix computed using the conjugate transpose of the key. Based on the rules of complex multiplication, the value embeddings are then aggregated using both real and imaginary attention weights to form the complex-valued attention output as follows.
\begin{equation}
\begin{aligned}
\text{Re}(\mathbf{Y}) &= \mathbf{A}_{\text{Re}} \cdot \text{Re}(\mathbf{V}) - \mathbf{A}_{\text{Im}} \cdot \text{Im}(\mathbf{V}), \\
\text{Im}(\mathbf{Y}) &= \mathbf{A}_{\text{Re}} \cdot \text{Im}(\mathbf{V}) + \mathbf{A}_{\text{Im}} \cdot \text{Re}(\mathbf{V}).
\end{aligned}
\label{eq:complex_attention_output}
\end{equation}

This allows the model to dynamically and selectively integrate statistical information and high-frequency details across both spatial and channel dimensions. Compared with simple concatenation or element-wise addition, this guided fusion strategy enables the network to more effectively focus on informative features and suppress irrelevant noise.




\textit{Loss function of the proposed method:}
To optimize the proposed complex denoising diffusion model, we employ the mean squared error (MSE) as the loss function, which corresponds to the commonly used \( L_2 \) loss. The \( L_2 \) loss encourages the model to produce predictions that are close to the ground truth in a Euclidean sense and is particularly effective in stabilizing training and accelerating convergence. Formally, the loss function at each time step \( t \) is defined as:
\begin{equation}\mathcal{L}_{\mathrm{MSE}}=\mathbb{E}_{\textbf{L}_0,\epsilon,t}\left[\|\epsilon-\hat{\epsilon}_\theta(\textbf{L}_t,\textbf{H}_1,t)\|_2^2\right]\end{equation}

\subsection{ High-frequency Detail Enhancement Module}
For an \( l \)-level contourlet transformation, high-frequency coefficients are obtained with \( l \) scales and \( 2^j \) directions. The challenge is to how to effectively and systematically integrate these multi-scale and multi-directional features. In this paper, we propose a high-frequency detail enhancement module, as illustrated in Fig.\ref{fig:fig1}(b). This module consists of two key components: cross-attention multi-directional enhancement within each scale and dilation convolution-based multi-scale fusion.

As shown in Fig.\ref{fig:fig1}(b), within each scale, each high-frequency subband captures structural information in a specific direction, and the different bands represent complementary directions. The main goal in the same scale is to combine multi-directional information to enhance diverse geometric features. So, we define the cross-attention enhancement module. First, we enhance the 8 directional coefficients (when \( j=3 \)) using 8 depth-wise separable convolutions, which efficiently learn individual features. Then, four cross-attention layers are employed,each adopting a similar attention structure as described in the fusion mechanism above, to leverage the different directional coefficients and obtain complementary information. Each pair of directional subbands is cross-fused, this process helps integrate the diverse directional details, improving the overall representation of edge and texture features for subsequent classification tasks. 

For different scales, a dilation convolution-based multi-scale fusion is designed to learn different scale information. Multi-level dilation convolutions are utilized to expand the receptive field and capture large-scale information. The detailed structure is composed of several dilation convolution layers, each progressively increasing the receptive field, allowing the model to capture both local and global features. Additionally, skip connections between the layers help preserve fine details from lower scales, ensuring that both high-level contextual and low-level textural features are effectively fused for enhanced classification performance.

\subsection{ Low- and High-frequency Feature Joint Classification}
Low and high-frequency feature joint classification combines both low-frequency (which typically captures textural patterns) and high-frequency (which captures edge and detail information) information together. By jointly leveraging both sets of features, this approach can improve classification accuracy, as it incorporates both statistical and structural image characteristics. To be specific, low-frequency features $\textbf{F}1$ from KCDM and high-frequency feature $\textbf{F}2$ are connected in channel dimension, and then several layer convolutions are employed to jointly learn them. Finally, a mapping layer and softmax layer are employed to obtain the classification map. The whole network learning is expressed as:
\begin{equation}
    \mathrm{\textbf{P}}=\mathrm{softmax}(\mathrm{MP}(\mathrm{CNN}(\textbf{F}1+\textbf{F}2)))
\end{equation}
where \textbf{P} is the network output. $\textbf{F}1$ and $\textbf{F}2$ are the features learned from KCDM in low-frequency and CAFE in high-frequency parts, respectively, defined as:
\begin{equation}
\begin{aligned}
  &F1 = \text{KCDM}(\textbf{L}1, \text{SKEM}(\textbf{H}1)) \\
  &F2 = \text{CAFE}(\textbf{H}1)
\end{aligned}
\end{equation}
where SKEM($\cdot$) is the module of structural knowledge extraction.
\textbf{L}1 and \textbf{H}1 are the low- and high- frequency coefficients from complex Contourlet transformation, expressed as: 
 \begin{equation}
  [\textbf{L}1,\textbf{H}1]=\text{ComContourlet(\textbf{X})} 
 \end{equation}
 
 Finally, the cross-entropy loss is utilized as the loss function, denoted by:
\begin{equation}Q=-\frac{1}{N}\sum_{i=1}^N\sum_{j=1}^Cy_{ij}\log\left(p_{ij}\right)\end{equation}
let \( N \) be the total number of samples, and \( y_{ij} \) denote the true label of sample \( i \). If sample \( i \) belongs to class \( j \), then \( y_{ij} = 1 \); otherwise, \( y_{ij} = 0 \). \( p_{ij} \) represents the probability that the model predicts sample \( i \) as belonging to class \( j \).The whole algorithm procedure of the proposed CD-KCDM is given in \textbf{Algorithm 1}.
\begin{algorithm*}[tb]
\small
	\caption{Algorithmic Procedure of the Proposed CD-KCDM Method}
	\label{Algorithm 1}
	\begin{algorithmic}
		\STATE {\bfseries Input:} PolSAR covariance matrix $\mathbf{C}$ and label map $L$.
		\STATE {\bfseries Output:} Classification map $\mathbf{Y}$.
		
		\emph{Step 1:} Apply the complex contourlet transform to the PolSAR matrix $\mathbf{C}$ to obtain the low-frequency and high-frequency subbands, denoted as $\mathbf{L1}$ and $\mathbf{H1}$, respectively.\\
		\emph{Step 2:} Feed the $\mathbf{L1}$ and the $\mathbf{H1}$ (enhanced by the Structural Knowledge Enhancement Module) into the KCDM to extract statistical features $\mathbf{F1}$.\\
        \emph{Step 3:} Enhance the $\mathbf{H1}$ using the High-frequency Detail Enhancement Module to obtain refined features $\mathbf{F2}$.\\
		\emph{Step 4:} Fuse $\mathbf{F1}$ and $\mathbf{F2}$, and perform dimensionality reduction via PCA.\\
		\emph{Step 5:} Classify the reduced features using a CNN-based classifier to generate the final prediction map $\mathbf{Y}$.\\

	\end{algorithmic}
\end{algorithm*}

\section{ EXPERIMENTS}
\subsection{Datasets}
Our experiments are conducted on three real PolSAR datasets collected from different frequency bands and sensors. Due to anonymity requirements, the details of the datasets and implementation codeS will be released in the final version. The three datasets are introduced below.

\textbf{Xi'an dataset:} This dataset consists of C-band polarimetric SAR (PolSAR) data acquired over Xi'an, China, using the RADARSAT-2 system. The image size is 512 $\times$ 512 pixels, with a spatial resolution of 8 $\times$ 8 meters. The scene is primarily composed of three land cover types: water, grass, and buildings. 

\textbf{Flevoland dataset:} Collected over Flevoland, the Netherlands, this dataset contains C-band PolSAR data with an image size of 1400 $\times$ 1200 pixels. It includes four major land cover classes: water, urban areas, woodland, and cropland.

\textbf{San Francisco dataset:}The dataset was acquired in 1985 over the San Francisco region using L-band PolSAR by NASA/JPL's AIRSAR system. It offers detailed characterization of complex terrain, with a resolution of 900 $\times$ 1024 pixels. Five primary land cover types are annotated: bare soil, mountains, ocean, urban areas, and vegetation.
\subsection{Experimental Setup}
The method is implemented using the PyTorch framework and trained on a system equipped with an Nvidia GeForce RTX 3060 GPU and an Intel Core i7-10700 CPU. The reverse process is configured with 16 time steps. To train the denoising U-Net, the learning rate is set to \(1 \times 10^{-4}\) and the patch size is fixed at 16. The model is optimized using Adam for 1000 iterations and select layer 1 for low-frequency feature extraction. For the classification module, 5\% of the dataset is used for training, while 1\% is reserved for validation. To ensure a fair comparison, all baseline methods are trained under the same experimental settings. Additionally, a series of ablation studies are conducted to provide a clear understanding of the contribution of each individual module. Finally, we analyze key parameters across different approaches. In the evaluation phase, we adopt Average Accuracy (AA), Overall Accuracy (OA), and the Kappa coefficient as performance metrics.
\subsection{Comparison with SOTA Methods}
To validate the effectiveness of the proposed method in the PolSAR image classification task, we compare it against five representative state-of-the-art approaches. The baseline methods include CVCNN \cite{RN50}, PolMPCNN \cite{RN14}, SGCN-CNN \cite{RN12}, SARDDPM \cite{RN49}, and SpectralDiff \cite{RN44}.

\begin{table}[htbp]
\centering
\caption{Results of different methods on Xi'an dataset(\%)}
\label{tab:512_table}
\resizebox{\linewidth}{!}{ 
\begin{tabular}{lcccccc}
\hline
\textbf{Class} & \textbf{CVCNN} & \textbf{PolMPCNN} & \textbf{SGCN-CNN} & \textbf{SARDDPM} & \textbf{SpectralDiff} & \textbf{OURS} \\
\hline
Water     & \textbf{97.26} & 95.52 & 86.57 & 84.37 & 89.70 & 90.89 \\
Grass     & 85.97 & 90.95 & 93.33 & 96.66 & 97.09 & 97.02 \\
Building  & 91.43 & 97.68 & 90.12 & 97.03 & 97.74 & \textbf{98.30} \\
\hline
OA        & 89.59 & 94.01 & 91.18 & 94.95 & 96.21 & \textbf{96.55}\\
AA        & 91.55 & 94.71 & 90.01 & 92.69 & 94.85 & \textbf{95.40} \\
Kappa     & 93.02 & 90.25 & 85.33 & 91.59 & 93.73 & \textbf{94.29} \\
\hline
\end{tabular}
}
\end{table}
\begin{table}[htbp]
\centering
\caption{Results of different methods on San Francisco dataset(\%)}
\label{tab:9010_table}
\resizebox{\linewidth}{!}{%
\begin{tabular}{lcccccc}
\hline
\textbf{Class} & \textbf{CVCNN} & \textbf{PolMPCNN} & \textbf{SGCN-CNN} & \textbf{SARDDPM} & \textbf{SpectralDiff} & \textbf{OURS} \\
\hline
Bare Soil   & 83.06 & 73.21 & 85.73 & 82.07 & 76.06 & \textbf{93.34} \\
Mountain    & 96.02 & 98.51 & 99.20 & 92.40 & 94.42 & \textbf{99.37} \\
Ocean       & 96.13 & 99.53 & 99.60 & 97.53 & 99.09 & \textbf{99.63} \\
Urban       & 94.25 & 99.73 & \textbf{99.81} & 97.63 & 98.45 & 99.77 \\
Vegetation  & 76.87 & 97.24 & 92.59 & 88.46 & 90.60 & \textbf{97.08} \\
\hline
OA          & 93.81 & 98.93 & 98.53 & 96.30 & 97.49 & \textbf{99.48} \\
AA          & 89.27 & 93.64 & 95.19 & 91.62 & 91.72 & \textbf{98.05} \\
Kappa       & 90.45 & 98.32 & 97.69 & 94.18 & 96.04 & \textbf{99.19} \\
\hline
\end{tabular}
}
\end{table}
\begin{table}[htbp]
\centering
\caption{Results of different methods on Flevoland dataset(\%)}
\label{tab:1412_table}
\resizebox{\linewidth}{!}{%
\begin{tabular}{lcccccc}
\hline
\textbf{Class} & \textbf{CVCNN} & \textbf{PolMPCNN} & \textbf{SGCN-CNN} & \textbf{SARDDPM} & \textbf{SpectralDiff} & \textbf{OURS} \\
\hline
Urban      & 96.26 & 96.29 & 98.56 & 69.33 & 92.76 & \textbf{99.39} \\
Water      & \textbf{99.85} & 99.14 & 98.31 & 81.86 & 98.05 & 99.67 \\
Woodland   & 96.48 & 98.77 & 98.25 & 75.76 & 96.82 & \textbf{99.31} \\
Cropland   & 93.94 & 98.66 & 98.39 & 84.54 & 96.84 & \textbf{99.41} \\
\hline
OA         & 96.57 & 98.49 & 98.35 & 79.24 & 96.59 & \textbf{99.45} \\
AA         & 96.63 & 98.21 & 98.38 & 77.87 & 96.12 & \textbf{99.44} \\
Kappa      & 95.33 & 97.94 & 97.75 & 71.59 & 95.35 & \textbf{99.25} \\
\hline
\end{tabular}
}
\end{table}

\begin{figure*}[htbp] 
    \centering 
    \includegraphics[scale=0.62]{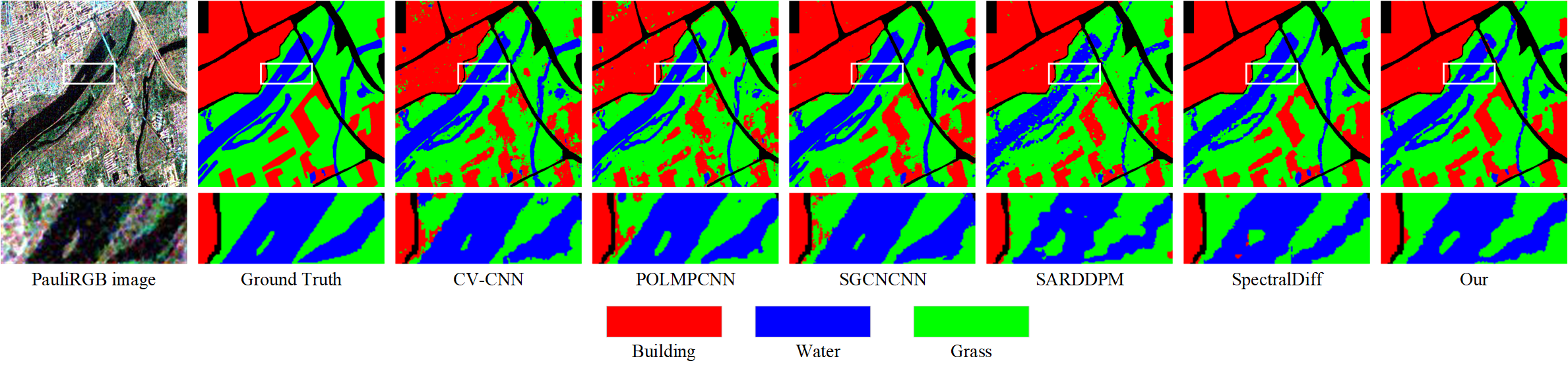} 
    \caption{Qualitative results comparison on the Xi'an dataset.} 
    \label{fig:512_fig} 
\end{figure*}

\begin{figure*} 
    \centering 
    \includegraphics[scale=0.42]{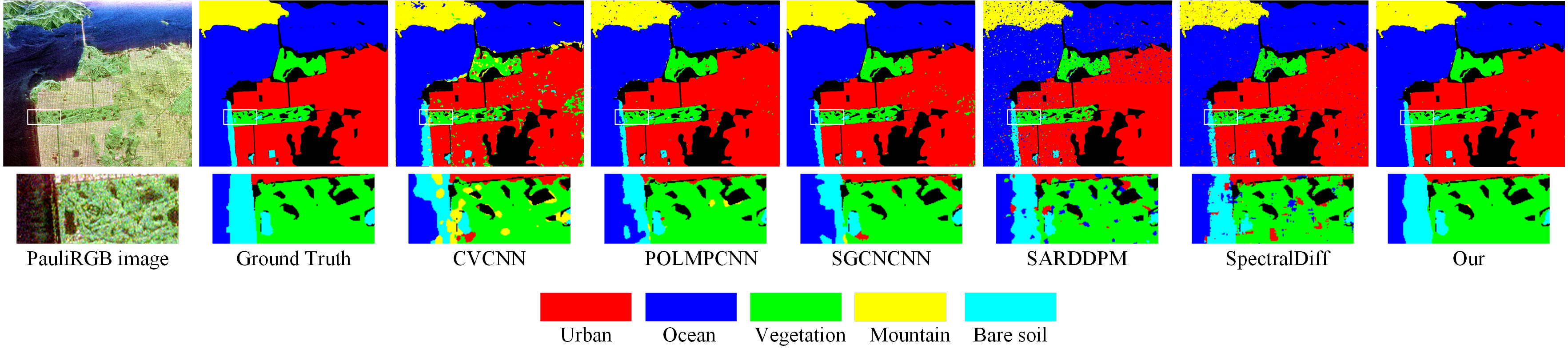} 
    \caption{Qualitative results comparison on the  San Francisco dataset.} 
    \label{fig:9010_fig} 
\end{figure*}
\begin{figure*}[htbp]
    \centering 
    \includegraphics[scale=0.6]{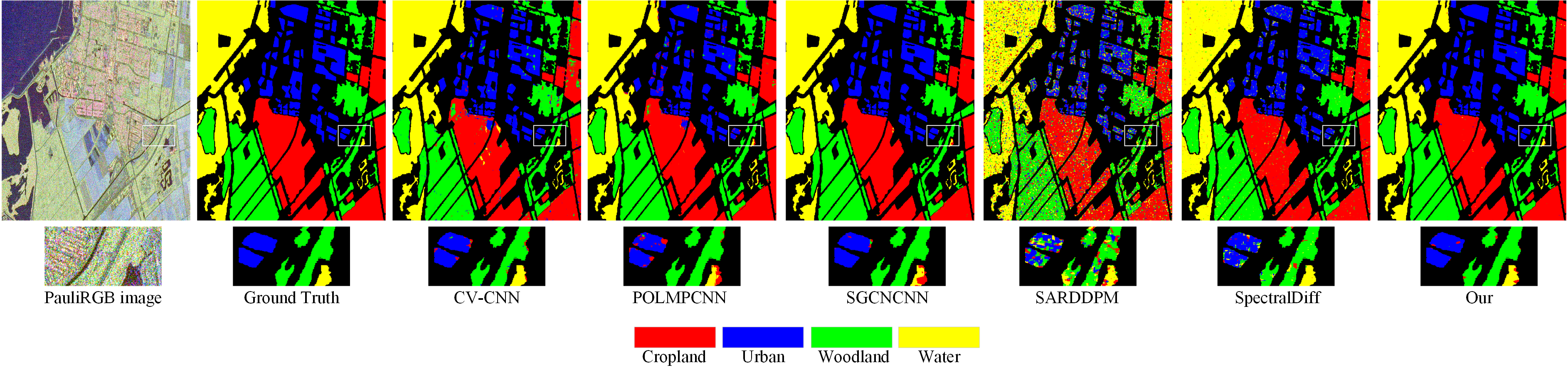} 
    \caption{Qualitative results comparison on the Flevoland dataset.} 
    \label{fig:1412_fig} 
\end{figure*}

\textit{1) Performance on the Xi'an Dataset:}
The quantitative and qualitative results are presented in Table \ref{tab:512_table} and Fig.\ref{fig:512_fig}, respectively. As can be seen in Table \ref{tab:512_table}, although CV-CNN and PolMPCNN achieve relatively high accuracy in the Water class, they tend to misclassify parts of the Grass region as Water and building, resulting in lower classification accuracy for the Grass category,this result is also presented in Fig.\ref{fig:512_fig}. SGCNCNN shows a more balantced performance across classes, but it fails to achieve the highest accuracy in any individual metric. In contrast, SARDDPM and SpectralDiff demonstrate noticeable improvements in OA, yet their performance on the Water class slightly declines due to the lack of structural guidance. The proposed HL-CDM method outperforms all baseline approaches across all three evaluation metrics . Specifically, compared to CVCNN, PolMPCNN, SGCN-CNN, SARDDPM, and SpectralDiff, the HL-CDM improves OA by 6.96\%, 2.54\%, 5.37\%, 1.60\%, and 0.34\%, respectively. These results confirm the robustness and effectiveness of the proposed method in accurately distinguishing various land cover types and improving the overall classification performance for PolSAR imagery.

\textit{2) Performance on the  San Francisco Dataset:}
Table \ref{tab:9010_table} and Fig.\ref{fig:9010_fig} presented the quantitative and qualitative results on the San Francisco dataset, respectively.As
depicted in Fig.\ref{fig:9010_fig},the CVCNN exhibits class
confusion between bare ground and mountains,POLMPCNN and SGCNCC have misclassification in terms of boundaries,it is evident that both SARDDPM and SpectralDiff suffer from noticeable speckle noise in the classification maps. This can be attributed to the excessive smoothing effect introduced during spatial-domain denoising, which leads to the loss of critical structural details. CV-CNN and PolMPCNN also exhibit scattered misclassification patches, particularly in Urban and Cropland regions, while SGCN-CNN produces relatively better visual quality. However, SGCN-CNN still struggles at category boundaries. In contrast, the proposed HL-CDM method demonstrates superior performance in preserving object boundaries and achieves the most visually coherent results among all methods. Specifically, the proposed method improves OA by 2.88\%, 0.96\%, 1.10\%, 20.21\%, and 2.86\% compared to CVCNN, PolMPCNN, SGCN-CNN, SARDDPM, and SpectralDiff, respectively.

\textit{3) Performance on the Flevoland Dataset:}
Quantitative performance metrics are reported in Table \ref{tab:1412_table}, while qualitative visualizations in Fig.\ref{fig:1412_fig} further validate the effectiveness of the proposed approach. As depicted in Fig. \ref{fig:1412_fig}, CV-CNN and POLMPCNN exhibit frequent misclassifications of urban areas as cropland. While SGCNCNN demonstrates improved performance in this regard, it tends to misclassify water bodies as cropland. Both SARDDPM and SpectralDiff present localized misclassification artifacts in certain regions.
The proposed HL-CDM method achieves the best visual results overall by effectively integrating structural and textural information to enhance class separability. In terms of overall accuracy, HL-CDM surpasses CVCNN, PolMPCNN, SGCN-CNN, SARDDPM, and SpectralDiff by 5.67\%, 0.55\%, 0.95\%, 3.18\%, and 1.99\%, respectively, further confirming its superior classification performance.

\subsection{Ablation Study}
To verify the effect of each proposed components and the parameters on the matching performance, we conduct ablation experiments on the Xi'an dataset.
\begin{table}[htbp]
\centering
\caption{Ablation Study of Key Modules on Xi'an dataset(\%)}
\label{tab:ablation}
\resizebox{\linewidth}{!}{%
\begin{tabular}{ccccccc}
\hline
No-NSCT & Low-Only & High-Only & Guided Fusion & OA & AA & Kappa \\
\hline
\checkmark &  &  &  & 96.35 & 95.69 & 93.98 \\
 & \checkmark &  &  & 93.85 & 92.68 & 89.82 \\
 &  & \checkmark &  & 95.20 & 94.12 & 92.06 \\
 & \checkmark & \checkmark &  & 96.26 & 94.74 & 93.80 \\
 & \checkmark & \checkmark & \checkmark & \textbf{96.55} & \textbf{95.40} & \textbf{94.29} \\
\hline
\end{tabular}
}
\end{table}

\textit{1) Effect of key Modules.}
The ablation study is conducted to evaluate the effectiveness of each component proposed in our method by isolating and testing them individually. Specifically, \textit{No-NSCT} denotes that the original data is directly processed by the diffusion model . \textit{Low-Only} uses only the low-frequency subbands, while \textit{High-Only} leverages only the enhanced high-frequency texture information. \textit{Guided Fusion} introduces semantic guidance by structurally enhancing and fusing the high-frequency information into the low-frequency representation classification.

As shown in Table~\ref{tab:ablation}, both \textit{LowF} and \textit{HighF} alone yield limited performance. The \textit{No-NSCT} setting performs better, as the diffusion model implicitly captures multi-frequency features. However, the best results are achieved when the guided fusion module is incorporated. By injecting high-frequency knowledge into the low-frequency stream, the model is able to better preserve fine-grained details and semantic boundaries. This confirms the complementary nature of frequency-aware decomposition and guided enhancement in our design.

\begin{figure}[htbp]
    \centering
    \includegraphics[width=0.3\textwidth]{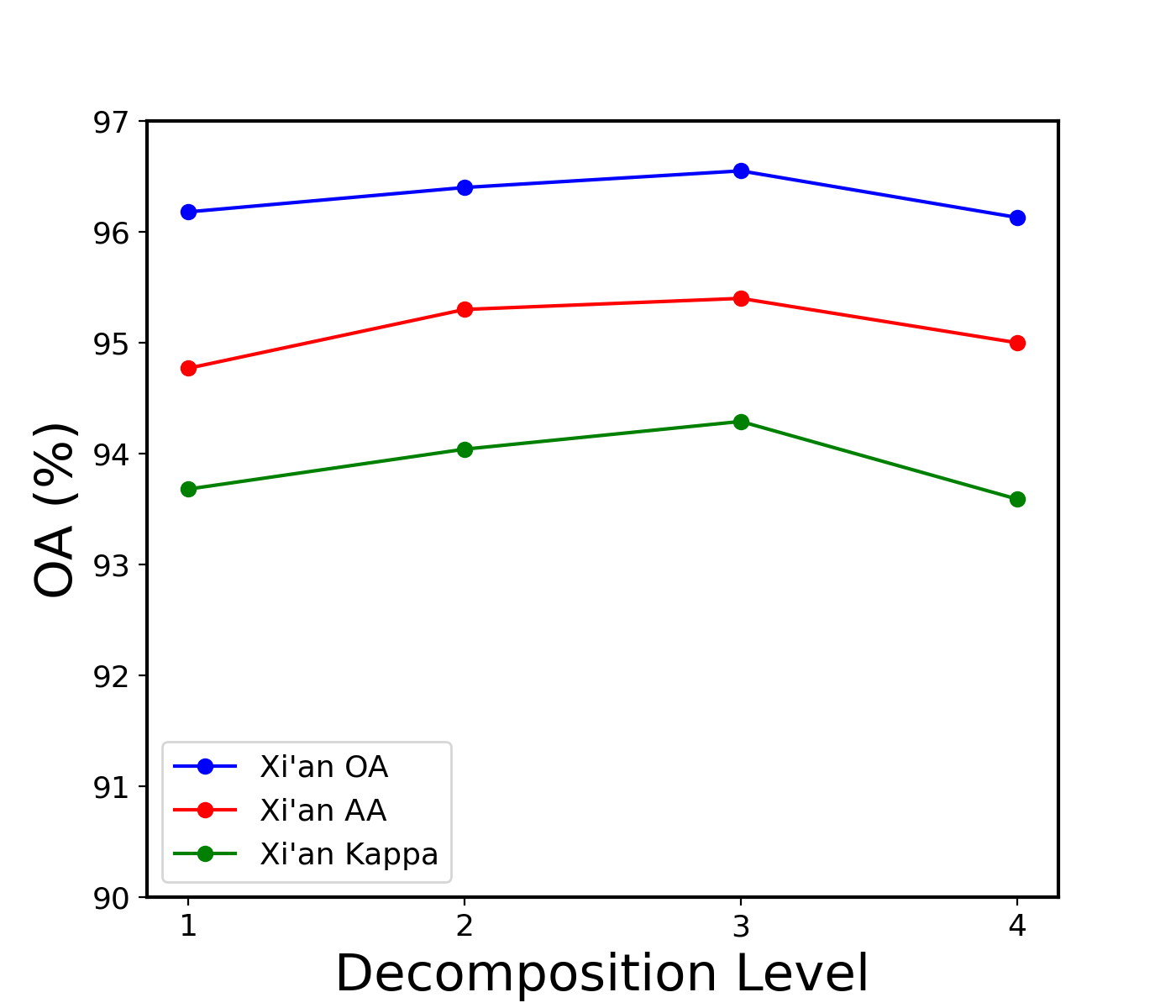}
    \caption{Performance variations with different NSCT decomposition levels on the Xi'an dataset.}
    \label{fig:decomp_levels}
\end{figure}

\textit{2) Effect of Contourlet Decomposition Levels.}
Fig.\ref{fig:decomp_levels} summarizes the impact of varying the number of NSCT decomposition levels on classification performance, as evaluated on the Xi'an dataset. As shown in Fig.\ref{fig:decomp_levels}, increasing the decomposition level initially leads to consistent improvements across all three evaluation metrics, peaking at level 3. However, when the decomposition level is further increased to 4, a noticeable performance drop is observed. This degradation is likely due to over-decomposition, where excessive splitting of high-frequency components leads to the loss of discriminative fine-grained features, ultimately impairing classification performance.

\textit{3) Effect of U-Net Depth on Feature Extraction.}
Table ~\ref{tab:U-NetLayer} illustrates the classification performance achieved with different numbers of layers in the U-Net feature extraction module. It can be seen that layer 1 can obtain superior performance in three indicators. The main reason is that it can obtain more suitable features from removing the noisy image to clean image in layer 1. With increasing network depth, feature representations become progressively more abstract and compressed, which can lead to the loss of fine-grained details, affecting classification accuracy.

\begin{table}[htbp]
    \centering
    \caption{Classification results with different U-Net layers on the Xi'an dataset(\%)}
    \label{tab:U-NetLayer}
    \setlength{\tabcolsep}{6mm}
    \begin{tabular}{cccc}
        \hline
        Layer & OA  & AA  & Kappa  \\
        \hline
        1 & \textbf{96.55} & \textbf{95.40} & \textbf{94.29} \\
        2 & 96.32 & 94.96 & 93.91 \\
        3 & 95.94 & 94.45 & 93.27 \\
        \hline
    \end{tabular}
\end{table}

\section{CONCLUSION}
In this paper, we have proposed a structural knowledge-guided complex diffusion model for PolSAR image classification, which effectively integrates multi-scale and multi-directional structural priors into a complex-valued diffusion framework. By employing the NSCT, the proposed method decomposes PolSAR data into low- and high-frequency subbands, allowing for dedicated processing of statistical and structural information.  A dual-stream network is designed, where the complex-valued diffusion model captures low-frequency statistical patterns, while enhanced high-frequency coefficients serve as structural guidance to refine spatial details and semantic boundaries. Experimental results on three sets of real PolSAR data demonstrate that the proposed HL-CDM can provide statistically higher accuracy than the state-of-the-art methods.

\section*{Acknowledgments}
This work was supported in part by the National Natural Science Foundation of China under Grant 62006186,62272383, the Youth Innovation Team Research Program Project of Education Department in Shaanxi Province under Grant 23JP111.


\vspace{11pt}

\vspace{-30pt}
\vfill

\end{document}